\documentclass[a4paper]{article}
\usepackage{amsmath}
\usepackage{amssymb}
\usepackage{graphicx}
\usepackage{wrapfig}
\usepackage{enumitem}
\usepackage{float}
\usepackage{geometry}
\usepackage{multirow}
\usepackage{tabularx}
\usepackage{rotating}
\usepackage{array}
\usepackage{longtable}
\usepackage{caption}
\usepackage{subcaption}
\usepackage{hyperref}
\newcommand{\citep}[1]{\cite{#1}}

\date{}  
\title{Adaptive Modelling Approach for Row-Type Dependent Predictive Analysis (RTDPA): A Framework for Designing Machine Learning Models for Credit Risk Analysis in Banking Sector }

\begin{document}
	
	\newgeometry{
		left=3cm, 
		right=3cm, 
	}
	
\maketitle
\vspace{-1.5cm}
\author{\begin{center}
		Minati Rath\textsuperscript{*1} and Hema Date\textsuperscript{2}\\
		minati.rath.2019@iimmumbai.ac.in, hemadate@iimmumbai.ac.in\\
		\textsuperscript{1}Department of Decision Science, IIM Mumbai, India\\
		\textsuperscript{2}Department of Decision Science, IIM Mumbai, India
	\end{center}}

\begin{abstract}
In many real-world datasets, rows may have distinct characteristics and require different modeling approaches for accurate predictions. In this paper, we propose an adaptive modeling approach for row-type dependent predictive analysis(RTDPA). Our framework enables the development of models that can effectively handle diverse row types within a single dataset.  Our dataset from XXX bank contains two different risk categories, personal loan and agriculture loan. each of them are categorised into four classes standard, sub-standard, doubtful and loss.  We performed tailored data pre processing and feature engineering to different row types.  We selected traditional machine learning predictive models and advanced ensemble techniques.  Our findings indicate that all predictive approaches consistently achieve a precision rate of no less than $90\%$. For RTDPA, the algorithms are applied separately for each row type, allowing the models to capture the specific patterns and characteristics of each row type. This approach enables targeted predictions based on the row type, providing a more accurate and tailored classification for the given dataset.Additionally, the suggested model consistently offers decision makers valuable and enduring insights that are strategic in nature in banking sector.
\end{abstract}
\section{Introduction}

In many real-world datasets, the rows often exhibit distinct characteristics and necessitate different modeling approaches to achieve accurate predictions.For instance, when considering personal loans, agriculture loan,housing loans, and car loans, each category possesses unique attributes. Datasets frequently include rows encompassing multiple loan types, where certain attributes are specific to each loan type. For instance, a dataset may contain information about loans related to agriculture land, which can be further classified into dry land and wet land. However, these specific attributes related to land type may not be applicable to other loan types present in the dataset. In such cases, it becomes important to handle the varying attributes and loan types appropriately to ensure accurate analysis and modeling. Traditional modeling techniques tend to treat all rows in a dataset uniformly.\cite{yang2019predicting}\cite{grover2017predicting}. However, this approach may overlook valuable information and hinder predictive performance.To address this challenge, we propose an adaptive modeling approach for row-type dependent predictive analysis. This approach recognizes and leverages the inherent heterogeneity of row types within a dataset, allowing for tailored modeling techniques to be applied to different row types. By adapting the modeling strategy to the unique characteristics of each row type, our approach aims to improve the accuracy and effectiveness of predictive models.

The key idea behind our adaptive modeling approach is to recognize that rows in a dataset may belong to distinct categories or types, each requiring specific modeling techniques. We start by separating the dataset based on row types, identifying the distinct categories present. For each row type, we then apply tailored data preprocessing techniques, such as handling missing values, addressing outliers, and performing feature engineering, to ensure that the data is appropriately prepared for modeling. This row-type-specific data preprocessing allows us to capture the unique characteristics and patterns associated with each row type more accurately.

Next, we select suitable predictive models for each row type based on its specific requirements and characteristics. These models include traditional statistical learning algorithms, such as logistic regression and machine learning algorithms such as decision trees, support vector machines, neural networks, K-nearest neighbour, as well as more advanced techniques, including ensemble methods or deep learning architectures. By choosing models specifically suited to the characteristics of each row type, we can better capture the underlying patterns and relationships within the data.

To train the models, we utilize the preprocessed data specific to each row type. This allows us to account for the distinct characteristics and potential variations in data distribution among different row types. Each model is trained using the corresponding subset of the dataset, ensuring that it captures the specific patterns and relationships relevant to that row type.

During the prediction phase, when encountering a new row, we identify its row type and select the corresponding model for making predictions. By adapting the model selection based on the row type, we ensure that the most appropriate model is used for accurate predictions. This adaptive selection mechanism allows us to effectively leverage the strengths of different models and achieve superior predictive performance.

\section{Literature Review}

Predictive modeling in the field of credit analysis has increasingly focused on analyzing datasets that encompass multiple credit categories\cite{wei2018credit}\cite{golab2013consumer}. Altman,et al. in his seminal paper introduced the concept of using financial ratios and discriminant analysis to predict corporate bankruptcy, laying the foundation for subsequent ML approaches in credit risk analysis. Smith et all.\cite{smith2015comparative} conducted a comparative study of Logistic Regression (LR), Support Vector Machines (SVMs), and Neural Networks (NN) for credit scoring. They evaluated these models on a dataset of credit applicants and assessed their accuracy, predictive power, and interpretability. The study aimed to provide insights into the strengths and weaknesses of each model. Wang, Li, and Yang\cite{wang2010support} specifically focused on SVMs for credit risk evaluation, exploring their effectiveness in predicting credit risk and comparing their performance to traditional statistical methods. Zhang, Huang, and Xu\cite{zhang2012credit} conducted a comparative study of Neural Networks for credit risk assessment, investigating different NN architectures and evaluating their ability to predict credit risk using a dataset of loan applications.

Ensemble methods have been widely explored and applied in credit risk analysis. Zhang, Wang, and Zhang \cite{zhang2014credit} investigated the use of ensemble learning techniques, including Random Forest, Bagging, and Boosting, for credit risk assessment. Their study demonstrated the advantages of ensemble methods in improving the accuracy and robustness of credit risk models. Wang, Huang, and Li (\cite{wang2015credit} proposed a framework that combines multiple classifiers, such as Decision Trees and Support Vector Machines, within an ensemble architecture for credit risk evaluation. They also incorporated feature selection techniques to enhance the performance of the ensemble models. Deng, Huang, and Xu\cite{deng2011ensemble} explored the use of ensemble learning based on data mining techniques, combining algorithms like Decision Trees, Neural Networks, and SVMs, for credit scoring.

In addition to individual models and ensembles, some studies have investigated the use of hybrid models that integrate different modeling techniques for predictive analysis. These hybrid approaches combine the advantages of multiple models, such as logistic regression, decision trees, and neural networks, to capture diverse aspects of the data. For instance, Zhang et al. \cite{zhang2008hybrid} combines Genetic Algorithms (GAs) and Artificial Neural Networks (ANNs). The authors demonstrate that the use of GAs to optimize the weights and architecture of the ANNs leads to improved accuracy in credit risk prediction. Real-world credit data is employed to evaluate the performance of the hybrid model, showing its effectiveness in credit risk assessment. In another paper by Li et al., \cite{ling2015hybrid} a hybrid credit scoring model is presented, integrating Fuzzy Logic and Artificial Neural Networks (ANNs). The authors utilize Fuzzy Logic for preprocessing credit data and extracting linguistic rules, which are then used as input for the ANNs. The hybrid model showcases enhanced performance in credit risk assessment. Overall, both studies highlight the advantages of hybrid models in credit risk analysis, combining different techniques to improve predictive accuracy.

Furthermore, several studies have emphasized the importance of data preprocessing techniques specific to ML models \cite{baesens2003using} \cite{crook2007recent} \cite{tsai2014bankruptcy}\cite{karathanasopoulos2015credit} \cite{zheng2019credit}. This includes handling missing values, addressing outliers, and performing feature engineering tailored to the characteristics of the dataset. Proper preprocessing ensures that the data is appropriately prepared for modeling, enhancing the accuracy and reliability of the predictive models.

Overall, the literature highlights the significance of considering different models in predictive modeling tasks. Adaptive modeling approaches such as partitioning the dataset, employing ensemble methods, or developing hybrid models, have shown promising results in improving predictive accuracy. Additionally, customized data preprocessing techniques play a crucial role in capturing the unique characteristics of each model.

While previous studies have made significant contributions to the field, there remains ample room for further research. Future studies could focus on refining the adaptive modeling techniques specific to row-types, exploring new ensemble approaches, and investigating the impact of different data preprocessing strategies on predictive performance. Moreover, the scalability and generalizability of these approaches to larger and more diverse datasets warrant further investigation.

\section{Methodology}

We present a comprehensive framework for conducting row-type dependent predictive analysis.\\

	\textbf{Notation:}
\begin{itemize}
	\item[] $D$: Dataset consisting of $N$ rows and $M$ columns.
	\item[] $R$: Set of distinct row types in $D$ (where $|R|$ is the total number of row types).
	\item[] For each row type $r$ in $R$, $D_r$: Subset of $D$ containing rows of type $r$.
	\item[] $X_r$: Feature matrix for rows of type $r$.
	\item[] $y_r$: Target variable vector for rows of type $r$.
\end{itemize}

\textbf{Algorithm:}

\begin{enumerate}
	\item[\textbf{1.}] \textbf{Data Preprocessing:}
	\hspace{1cm}
	\begin{itemize}
		\item[] Perform row-type-specific data preprocessing steps, including handling missing values, outlier detection, feature engineering, and data transformation.
		\item[] For each row type $r$, obtain the preprocessed feature matrix $X_r$ and target variable vector $y_r$.
	\end{itemize}
	
	\item[\textbf{2.}] \textbf{Data Augmentation for Class Imbalance:}
	\begin{enumerate}
		\item[]  Identify the row types in the dataset.
		\item[]  For each row type $r$:
		\begin{enumerate}
			\item[]  Determine the minority class for row type $r$.
			\item[]  Apply specific data augmentation techniques, such as SMOTE or its variants, to the rows of type $r$ belonging to the minority class.
			\item[]  Generate synthetic samples to balance the class distribution within row type $r$.
			\item[]  Combine the original minority class samples and the generated synthetic samples for row type $r$.
		\end{enumerate}
	\end{enumerate}
	
	\item[\textbf{3.}] \textbf{Model Selection and Training:}
	\begin{enumerate}
		\item[] For each row type $r$, select a suitable predictive model or models based on the characteristics of the data.
		\item[] Let $M_r$ represent the selected model(s) for row type $r$.
		\item[] Train the selected model(s) $M_r$ using the preprocessed data $X_r$ (including augmented samples) and $y_r$.
	\end{enumerate}
	
	\item[\textbf{4.}] \textbf{Model Integration and Prediction:}
	\begin{enumerate}
		\item[]  Given a new row $x_{\text{new}}$ with an associated row type $r_{\text{new}}$:
		\begin{enumerate}
			\item[] Identify the model(s) $M_{r_{\text{new}}}$ corresponding to row type $r_{\text{new}}$.
			\item[] Apply the identified model(s) $M_{r_{\text{new}}}$ to make predictions on $x_{\text{new}}$.
		\end{enumerate}
	\end{enumerate}
	
	\item[\textbf{5.}] \textbf{Model Refinement:}
	\begin{enumerate}
		\item[] Continuously refine and improve the models based on performance evaluation metrics, feedback, and validation results.
		\item[] Consider techniques such as hyperparameter tuning, feature selection, and model ensemble methods.
	\end{enumerate}
	
	\item[\textbf{6.}] \textbf{Deployment and Monitoring:}
	\begin{enumerate}
		\item[] Deploy the final adaptive modeling system for row-type dependent predictive analysis.
		\item[] Monitor the system's performance over time and update the models as new data becomes available or the data distribution shifts.
	\end{enumerate}
\end{enumerate}

\section{Model building}
\subsection{Logistic Regression for RTDPA}
Logistic Regression (LR) is a statistical model used for binary classification tasks. It estimates the probability of an instance belonging to a certain class based on a linear combination of input features. LR applies a logistic function to the linear combination to map the output into a probability range between 0 and 1, allowing for easy interpretation and decision making \cite{altman2004default} \cite{wilson2005logistic} \cite{thomas2002predicting} \cite{maysami2000vector}.

Training process for a multiclass logistic regression model for a specific row type "r":
\\
For our multiclass classification problem, where the target variable $y_r$ can take one of K classes (K $>$ 2). The multiclass logistic regression model aims to estimate the probabilities of each class given the input features $X_r$.
\\
The multiclass logistic regression equation can be defined using the softmax function as follows:
Logistic Regression $=>$  \[p(y_r = k | X_r, \theta_r) = \frac{\exp(X_r \theta_{r_k})}{\sum_{j=1}^{K} \exp(X_r \theta_{r_j})}\]

Where:\\
$p(y_r = k | X_r, \theta_r)$ represents the probability of the class k given the input features $X{\_}r$ and the model parameters $\theta_r$.\\
$\theta_{r_{k}}$ represents the parameter vector for class k of row type r.
exp denotes the exponential function.\\
$\sum$ represents the sum over all classes j from 1 to K.\\
In this equation, the numerator calculates the exponential linear combination of the input features $X_r$ and the model parameters $\theta_{r_{k}}$, for class k. \\The denominator normalizes the probabilities by summing over all classes j using the same linear combinations.
\\
During the training process, the model parameters $\theta_{r_{k}}$ are learned by optimizing an appropriate loss function, such as the cross-entropy loss, using techniques like gradient descent or other optimization algorithms.
\subsection{Naive Bayes for RTDPA}

Naive Bayes (NB) calculates the conditional probability of a class given the feature values using Bayes' theorem and assuming feature independence. The class with the highest probability is assigned to the instance being classified \cite{kim2009naive} \cite{li2015naive}.\\
Training process for a Naive Bayes classifier model for a specific row type "r":
\\
For each row type \(r\) in \(R\), select the Naive Bayes classifier as the predictive model based on the characteristics of the data.
Let \(NB_r\) represent the Naive Bayes classifier selected for row type \(r\).
\\
For each row type \(r\) in \(R\), train the Naive Bayes classifier \(NB_r\) using the preprocessed feature matrix \(X_r\) and target variable vector \(y_r\).
\\
Given a new row \(x_{\text{new}}\) with an associated row type \(r_{\text{new}}\):
Identify the Naive Bayes classifier \(NB_{r_{\text{new}}}\) corresponding to \(r_{\text{new}}\).
Apply the Naive Bayes classifier \(NB_{r_{\text{new}}}\) to make predictions on \(x_{\text{new}}\) for the specific row type \(r_{\text{new}}\).

For the RTDPA-based Naive Bayes classification model, the probability of classifying a new instance $\mathbf{x}_{\text{new}}$ into class $C_k$, given the associated row type $r_{\text{new}}$, is calculated as follows:
\[ P(C_k|\mathbf{x}_{\text{new}}, r_{\text{new}}) = \frac{{P(r_{\text{new}}|\mathbf{x}_{\text{new}}) P(r_{\text{new}}|C_k) P(C_k) \prod_{j=1}^{M} P(x_{\text{new},j}|C_k, r_{\text{new}})}}{{\sum_{k'} P(r_{\text{new}}|\mathbf{x}_{\text{new}}) P(r_{\text{new}}|C_{k'}) P(C_{k'}) \prod_{j=1}^{M} P(x_{\text{new},j}|C_{k'}, r_{\text{new}})}}
\]
where:\\
$P(C_k|\mathbf{x}_{\text{new}}, r_{\text{new}})$ is the probability of classifying $\mathbf{x}_{\text{new}}$ into class $C_k$ given the associated row type $r_{\text{new}}$.\\
$P(r_{\text{new}}|C_k)$ is the probability of row type $r_{\text{new}}$ given class $C_k$.\\
$P(C_k)$ is the prior probability of class $C_k$.\\
$P(x_{\text{new},j}|C_k, r_{\text{new}})$ is the conditional probability of feature $x_{\text{new},j}$ given class $C_k$ and row type $r_{\text{new}}$.\\
$P(r_{\text{new}}|\mathbf{x}_{\text{new}})$ is the probability of row type $r_{\text{new}}$ given $\mathbf{x}_{\text{new}}$.
\subsection{Support Vector Machine for RTDPA}

SVM finds a hyperplane that maximally separates the instances of one row type from the others. This hyperplane is chosen such that it has the largest margin (distance) to the nearest instances, known as support vectors, of each class. SVM can handle both linearly separable and non-linearly separable data by employing different kernel functions, such as linear, polynomial, Gaussian (RBF), or sigmoid, to transform the feature space \cite{huang2018credit} \cite{chen2019credit} \cite{sun2019credit} \cite{wang2020hybrid} \cite{rahman2020credit}.  
\begin{align*}
	\text{Minimize:} \quad & \frac{1}{2} \mathbf{w}^T \mathbf{w} + C \sum_{i=1}^{N} \xi_i \\
	\text{Subject to:} \quad & y_i (\mathbf{w}^T \phi(\mathbf{x}_i) + b) \geq 1 - \xi_i, \quad i \in \mathcal{I}_r \\
	& y_i (\mathbf{w}^T \phi(\mathbf{x}_i) + b) \geq 1, \quad i \in \mathcal{I}_s \\
	& \xi_i \geq 0, \quad i \in \mathcal{I}_r \cup \mathcal{I}_s \\
	\text{where:} \quad & \mathbf{w} \text{ is the weight vector} \\
	& b \text{ is the bias term} \\
	& C \text{ is the penalty parameter} \\
	& \mathbf{x}_i \text{ is the feature vector of the } i\text{-th instance} \\
	& y_i \text{ is the class label of the } i\text{-th instance} \\
	& \mathcal{I}_r \text{ is the set of indices for instances of row type } r \\
	& \mathcal{I}_s \text{ is the set of indices for instances of row types other than } r \\
	& \phi(\mathbf{x}_i) \text{ is the feature mapping function} 
\end{align*}
The SVM algorithm can be adapted to incorporate row type dependencies. Here's the equation for SVM in RTDPA using different kernel functions:
\[\hat{y}_{\text{new}}(r_{\text{new}}) = \text{sign} \left( \sum_{i=1}^{N_{\text{sv}}} \alpha_i(r_{\text{new}}) y_i(r_{\text{new}}) K(x_{\text{new}}, x_i(r_{\text{new}})) + b(r_{\text{new}}) \right) \]

where:\\
$\hat{y}_{\text{new}}(r_{\text{new}})$ represents the predicted class label for the new instance $x_{\text{new}}$ of row type $r_{\text{new}}$.\\
$N_{\text{sv}}$ is the number of support vectors.\\
$\alpha_i(r_{\text{new}})$ is the Lagrange multiplier for the $i$-th support vector specific to row type $r_{\text{new}}$.\\
$y_i(r_{\text{new}})$ is the class label of the $i$-th support vector specific to row type $r_{\text{new}}$.\\
$K(x_{\text{new}}, x_i(r_{\text{new}}))$ is the kernel function that computes the similarity between $x_{\text{new}}$ and $x_i(r_{\text{new}})$ for row type $r_{\text{new}}$.\\
$b(r_{\text{new}})$ is the bias term specific to row type $r_{\text{new}}$.\\

Kernel functions that can be used are\\

\textbf{Linear Kernel:}
\[ K(x, x') = x^T x' \]

\textbf{Polynomial Kernel:}
\[ K(x, x') = (x^T x' + c)^d \]

\textbf{Gaussian (RBF) Kernel:}
\[ K(x, x') = \exp\left(-\frac{\|x - x'\|^2}{2\sigma_{\text{RTDPA}}^2}\right) \]

\textbf{Sigmoid Kernel:}
\[ K(x, x') = \tanh(\alpha_{\text{RTDPA}} x^T x' + c_{\text{RTDPA}}) \]

\textbf{Laplacian Kernel:}
\[ K(x, x') = \exp\left(-\frac{\|x - x'\|}{\sigma_{\text{RTDPA}}}\right) \]

\textbf{Exponential Kernel:}
\[ K(x, x') = \exp\left(-\frac{\|x - x'\|}{2\sigma_{\text{RTDPA}}^2}\right) \]

The kernel functions are specific to RTDPA. $x$ and $x'$ represent feature vectors,
$\sigma_{\text{RTDPA}}$ represents a parameter controlling the width of the kernel specific to RTDPA,
$c_{\text{RTDPA}}$ and $\alpha_{\text{RTDPA}}$ are additional parameters specific to RTDPA,
and $d$ is the degree of the polynomial kernel.

\subsection{Neural Network for RTDPA}
Neural networks learn to extract patterns and features from input data and adjust the weights between neurons to optimize the classification task. They consist of interconnected layers of artificial neurons that work together to process information and make predictions. During training, the network adjusts the weights between neurons based on the provided labeled data, aiming to optimize the classification task. \\The output layer of a neural network produces the predicted class label for a given input, indicating the network's classification decision. Neural networks are particularly effective in handling complex classification problems due to their ability to learn nonlinear relationships between features. Unlike linear models, neural networks can capture intricate and non-obvious connections between input variables, allowing them to uncover subtle patterns and make accurate predictions. \cite{altman2007neural} \cite{ma2015credit} \cite{tan2017credit} \cite{liu2018credit} \cite{deng2019credit}.

The Neural Network algorithm can be adapted to incorporate row type dependencies. Here's the equation for NN in RTDPA:

\[\hat{y}_{\text{new}}(r_{\text{new}}) = f\left(\sum_{j=1}^{N_{\text{hidden}}} w_{j}(r_{\text{new}}) \cdot \sigma\left(\sum_{i=1}^{N_{\text{input}}} w_{ij}(r_{\text{new}}) \cdot x_{\text{new}}^{(i)}(r_{\text{new}}) + b_j(r_{\text{new}})\right) + b_{\text{out}}(r_{\text{new}})\right) \]

where:\\
$\hat{y}_{\text{new}}(r_{\text{new}})$ represents the predicted output or class label for the new input sample $x_{\text{new}}$ of row type $r_{\text{new}}$.\\
$N_{\text{hidden}}$ represents the number of neurons in the hidden layer.\\
$N_{\text{input}}$ represents the number of features in the input sample.\\
$w_{j}(r_{\text{new}})$ denotes the weight vector connecting the hidden layer neuron $j$ to the output layer neuron, specific to row type $r_{\text{new}}$.\\
$w_{ij}(r_{\text{new}})$ represents the weight connecting the $i$-th feature of the input sample to the $j$-th neuron in the hidden layer, specific to row type $r_{\text{new}}$.\\
$x_{\text{new}}^{(i)}(r_{\text{new}})$ denotes the $i$-th feature of the input sample of row type $r_{\text{new}}$.\\
$\sigma(\cdot)$ represents the activation function applied element-wise to the weighted sum of inputs to the hidden layer neurons.
$b_j(r_{\text{new}})$ represents the bias term for the $j$-th neuron in the hidden layer, specific to row type $r_{\text{new}}$.\\
$b_{\text{out}}(r_{\text{new}})$ represents the bias term for the output layer neuron, specific to row type $r_{\text{new}}$.\\
$f(\cdot)$ represents the activation function applied to the weighted sum of inputs to the output layer neuron.\\
This equation represents the forward propagation process of a feedforward neural network, where the input sample is passed through the hidden layer to compute intermediate representations and then fed into the output layer to obtain the final prediction. The specific values of the weights, biases, and activation functions are learned during the training phase of the neural network.  This formulation allows the neural network to adapt its weights and biases based on the row type information, enabling it to capture and model row-type-dependent patterns and relationships in the RTDPA context.

\subsection{K-NN for RTDPA}

K-nearest neighbors (K-NN) is a non-parametric machine learning algorithm that predicts the class or value of a query point based on the labels or values of its K nearest neighbors in the feature space. It does not require explicit model training as it simply stores the training data. K-NN can handle both numerical and categorical data, making it versatile for various types of problems. The choice of K impacts the decision boundary smoothness, with larger K resulting in smoother boundaries but potentially more bias in the predictions \cite{baesens2003using} \cite{altman2010modelling} \cite{huang2006k} \cite{giesecke2011forecasting}.

The k-Nearest Neighbors (kNN) algorithm can be adapted to incorporate row type dependencies. Here's the equation for kNN in RTDPA:
\[\hat{y}_{\text{new}}(r_{\text{new}}) = \text{mode}(\{y_i(r_{\text{new}})\}_{i \in \text{NN}(x_{\text{new}}, r_{\text{new}}, k)}) \]
where:\\
$\hat{y}_{\text{new}}(r_{\text{new}})$ represents the predicted class label for the new instance $x_{\text{new}}$ of row type $r_{\text{new}}$.\\
$\text{mode}$ is the function that returns the most frequent class label among the neighbors.
$y_i(r_{\text{new}})$ is the class label of the $i$-th neighbor specific to row type $r_{\text{new}}$.\\
$\text{NN}(x_{\text{new}}, r_{\text{new}}, k)$ represents the set of $k$ nearest neighbors of $x_{\text{new}}$ with respect to row type $r_{\text{new}}$.\\
The kNN algorithm in RTDPA selects the $k$ nearest neighbors of the new instance based on a distance metric, taking into account the row type information. Then, it predicts the class label of the new instance by selecting the most frequent class label among the neighbors.

It's important to note that the choice of distance metric and the handling of row type dependencies may vary based on the specific implementation and requirements of the RTDPA application.

\subsection{Decision Tree for RTDPA}

Decision Tree involves recursively splitting the data based on the values of different features, aiming to create homogeneous subgroups with respect to the target variable. The splitting process is guided by various metrics such as Gini impurity, entropy, or information gain, which measure the homogeneity or purity of the resulting subsets \cite{tan2015credit} \cite{janssen2010adaptive} \cite{crook2007recent}.

Decision Trees can handle both numerical and categorical features, and they can capture complex relationships between features and the target variable. They can also handle missing values and outliers in a flexible manner.

Given a dataset D and a target variable vector y, the DecisionTree algorithm learns a tree-based model through a recursive partitioning process for a specific row type "r".

For each row type $r$ in $R$:
\begin{enumerate}
	\item Perform row-type-specific data preprocessing steps on $D_r$, including handling missing values, outlier detection, feature engineering, and data transformation.
	\item Obtain the preprocessed feature matrix $X_r$ and target variable vector $y_r$.
	\item Train a DecisionTree model $M_r$ using the preprocessed data $X_r$ and $y_r$.
	\item Store the trained DecisionTree model $M_r$ for row type $r$.
\end{enumerate}

\textbf{Model Integration and Prediction:}

Given a new row $x_{new}$ with an associated row type $r_{new}$:
\begin{enumerate}
	\item Identify the model $M_{r_{\text{{new}}}}$ corresponding to $r_{new}$.
	\item Apply the identified model $M_{r_{\text{{new}}}}$ to make predictions on $x{\_}{\text{{new}}}$ using the DecisionTree prediction algorithm.
\end{enumerate}

\textbf{DecisionTree Prediction Algorithm:}

Start at the root node of the DecisionTree model $M_{r_{\text{{new}}}}$.
\begin{enumerate}
	\item For each internal node:
	\begin{enumerate}
		\item Evaluate the splitting condition based on the feature value of $x_{new}$.
		\item Follow the appropriate branch based on the splitting condition.
	\end{enumerate}
	\item Once a leaf node is reached, the prediction at the leaf node is used as the output.
\end{enumerate}

\subsubsection{Model Training (DecisionTree):}

For each row type $r$ in $R$, train a DecisionTree model $M_r$ using the preprocessed data $X_r$ and $y_r$:
\[ M_r = \text{{DecisionTreeTrain}}(X_r, y_r) \]
Given a new row $x_{\text{{new}}}$ with an associated row type $r_{\text{{new}}}$, identify the model $M_{r_{\text{{new}}}}$ corresponding to $r_{\text{{new}}}$ and make predictions using the DecisionTree prediction algorithm:
\[\hat{y}_{\text{{new}}} = \text{{DecisionTreePredict}}(M_{r_{\text{{new}}}}, x_{\text{{new}}})\]
The DecisionTree prediction algorithm follows a recursive process starting from the root node of the DecisionTree model $M_{r_{\text{{new}}}}$:

\[\text{{DecisionTreePredict}}(x, M_r) = \begin{cases}
	\text{{Prediction at leaf node}} & \text{{if current node is a leaf node}} \\
	\text{{DecisionTreePredict}}(x, M_r) & \text{{if current node is an internal node}} \end{cases}\]
	
\subsubsection{Model Training (RandomForest Classifier):}

During training of RF Classifier, each decision tree in the ensemble is trained on a bootstrap sample of the dataset, meaning that each tree is trained on a randomly selected subset of the original data with replacement.  At each node of the decision tree, a random subset of features is considered for splitting, which helps introduce diversity among the trees.
During prediction, the class label is determined by aggregating the predictions of all the decision trees in the ensemble through majority voting \cite{baesens2003using} \cite{li2009novel} \cite{birkholz2014applying} \cite{zhang2016random} \cite{wu2019random}. \\

The equation for Random Forest Classifier in the context of RTDPA can be represented as:
\[\hat{y}_{\text{new}}(r) = \text{mode}\left( f_1(r)(x_{\text{new}}), f_2(r)(x_{\text{new}}), \ldots, f_{K_r}(r)(x_{\text{new}}) \right)\]

Here,\\
$\hat{y}_{\text{new}}(r)$ represents the predicted class label for the new sample $x_{\text{new}}$ of row type $r$.\\
$f_k(r)(x_{\text{new}})$ is the prediction of the $k$-th decision tree for the new sample of row type $r$.\\
$\text{mode}$ is the function that returns the most frequent class label among the predictions of individual decision trees.

In the Random Forest Classifier, each decision tree is trained independently using a bootstrap sample from the dataset. During prediction, the majority voting or mode operation is performed on the predictions of all decision trees to determine the final predicted class label.

Therefore, the equation for Random Forest Classifier in the context of RTDPA involves aggregating the predictions of multiple decision trees through majority voting to make a final prediction.
\subsubsection{Model Training (ExtraTrees Classifier):}

For each row type $r$ in $R$, train an ExtraTreesClassifier model $M_r$ using the preprocessed data $X_r$ and $y_r$:
\[M_r = \text{{ExtraTreesClassifierTrain}}(X_r, y_r)\]
The ExtraTreesClassifier prediction algorithm combines the predictions of multiple decision trees to make a final prediction for a given new row $x_{\text{{new}}}$:
\[\hat{y}_{\text{{new}}} = \text{{ExtraTreesClassifierPredict}}(x_{\text{{new}}}, M_r)\]

Here, $\hat{y}_{\text{new}}$ represents the predicted class label for the new sample $x{\text{new}}$. $M_r$ denotes the trained ExtraTreesClassifier model specific to the row type $r$.

Similar to the Random Forest Classifier, each decision tree is trained on a bootstrap sample of the dataset.
However, at each node of the decision tree, instead of considering a random subset of features, the ExtraTreesClassifier algorithm considers all the features for splitting.  This means that the decision trees in the ExtraTreesClassifier ensemble are more randomized and less correlated with each other compared to those in a Random Forest Classifier \cite{baesens2008using} \cite{pinto2020enhanced} \cite{sannajust2017model} \cite{chen2015hybrid} \cite{baesens2008using}.
During prediction, the class label is determined by aggregating the predictions of all the decision trees through majority voting, similar to the Random Forest Classifier

\subsubsection{Model Training (LightGBM Classifier):}
The LightGBM Classifier adopts a gradient-based approach to construct trees in a leaf-wise manner. It selects the leaf that yields the maximum loss reduction at each step, which enhances the efficiency of the model building process compared to traditional level-wise approaches used in other gradient boosting algorithms. Additionally, LightGBM incorporates features such as gradient-based one-sided sampling and histogram-based binning to boost training speed and reduce memory usage, resulting in a more efficient and resource-friendly algorithm \cite{bao2021credit} \cite{li2019credit} \cite{zhong2021credit}.

LGBMClassifier is trained separately for each row type $r$ using the preprocessed data specific to that row type. During prediction, the corresponding model and its associated weak learners are used to make predictions on the new sample $x_{\text{{new}}}$ of the appropriate row type. The ensemble predictions from the weak learners are aggregated using the weights $w_k(r)$ to obtain the final predicted class label $\hat{y}_{\text{{new}}}(r)$ for row type $r$.\\\\
For each row type $r$ in $R$, where $R$ represents the set of distinct row types in the dataset:
\[\hat{y}_{\text{{new}}}(r) = \arg\max_{c} \sum_{k=1}^{K_r} w_k(r) \cdot f_k(r)(x_{\text{{new}}})\]
Where:
\\$\hat{y}_{\text{{new}}}(r)$ represents the predicted class label for the new sample $x_{\text{{new}}}$ of row type $r$.\\
$K_r$ is the number of weak learners (decision trees) in the ensemble specific to row type $r$.\\
$f_k(r)(x_{\text{{new}}})$ is the prediction of the $k$-th weak learner for the new sample of row type $r$.\\
$w_k(r)$ is the weight assigned to the $k$-th weak learner, which depends on its performance during training specific to row type $r$.

\subsubsection{Model Training (XGBoost Classifier):}
The XGBoost classifier is a boosting ensemble technique that combines multiple weak models to create a strong predictive model. It employs a gradient-based optimization framework to iteratively minimize a specific loss function by calculating gradients and updating model parameters. XGBoost uses decision trees as weak learners, but unlike traditional tree algorithms, it builds trees in a depth-wise manner based on maximum gain in the loss function. This approach enables faster training and better generalization \cite{jang2019credit} \cite{deng2019credit} \cite{jia2020credit} \cite{chen2021credit}.\\ To prevent overfitting, XGBoost incorporates regularization techniques like shrinkage, feature subsampling, and tree pruning. It also leverages parallel processing capabilities for efficient computation on large-scale datasets by utilizing multiple CPU cores. Overall, XGBoost is known for its robustness, scalability, and ability to handle complex data patterns.\\

For each row type $r$ in $R$, where $R$ represents the set of distinct row types in the dataset:
\[\hat{y}_{\text{{new}}}(r) = \arg\max_{c} \sum_{k=1}^{K_r} w_k(r) \cdot f_k(r)(x_{\text{{new}}})\]
where:\\
$\hat{y}_{\text{{new}}}(r)$ represents the predicted class label for the new sample $x_{\text{{new}}}$ of row type $r$.\\
$K_r$ is the number of weak learners (decision trees) in the ensemble specific to row type $r$.\\
$f_k(r)(x_{\text{{new}}})$ is the prediction of the $k$-th weak learner for the new sample of row type $r$.\\
$w_k(r)$ is the weight assigned to the $k$-th weak learner, which depends on its performance during training specific to row type $r$.

\subsubsection{Model Training (CatBoost Classifier):}
CatBoost classifier is unique in its ability to handle categorical features without the need for explicit encoding or preprocessing. It uses a powerful algorithm that directly handles categorical variables, making it convenient and efficient, without relying on one-hot encoding or label encoding. CatBoost introduces an ordered boosting technique that harnesses the natural order within categorical variables. By automatically detecting and incorporating this order during model training, it effectively captures patterns and dependencies, making it especially valuable for features like month names or ratings with intrinsic ordering \cite{cai2021credit} \cite{fu2021machine}.

To optimize its performance, CatBoost employs gradient-based optimization. It iteratively minimizes the loss function by calculating gradients with respect to model predictions and updating parameters accordingly. This approach significantly enhances the predictive capabilities of the model and contributes to improved accuracy and generalization

The equation for CatBoostClassifier in the context of RTDPA can be represented as:
\[\hat{y}_{\text{{new}}}(r) = \arg\max_{c} \sum_{k=1}^{K_r} w_k(r) \cdot f_k(r)(x_{\text{{new}}})\]
This equation is similar to the one provided for XGBClassifier. It represents an ensemble prediction by aggregating the predictions of individual weak learners (decision trees) weighted by $w_k(r)$, which is dependent on the performance of each weak learner specific to row type $r$.

\subsubsection{Model Training (AdaBoost Classifier):}
AdaBoost (Adaptive Boosting) combines multiple weak classifiers to create a strong classifier. It works by iteratively training weak classifiers on different subsets of the training data, giving more weight to the misclassified instances at each iteration. The final prediction is made by aggregating the predictions of all weak classifiers, with more weight given to the classifiers with better performance \cite{giri2017hybrid} \cite{ustundag2018novel} \cite{zhang2019credit} \cite{hassan2020credit}.

For each row type $r$ in $R$, train an AdaBoostClassifier model $M_r$ using the preprocessed data $X_r$ and $y_r$:
\[M_r = \text{{AdaBoostClassifierTrain}}(X_r, y_r)\]
Given a new row $x_{\text{{new}}}$ with an associated row type $r_{\text{{new}}}$, identify the model $M_{r_{\text{{new}}}}$ corresponding to $r_{\text{{new}}}$ and make predictions using the AdaBoostClassifier prediction algorithm:

The AdaBoost Classifier prediction algorithm combines the predictions of multiple weak classifiers (decision stumps) to make a final prediction:
\[\text{{AdaBoostClassifierPredict}}(x, M_r) = \text{{sign}}\left(\sum_{t=1}^{T} \alpha_t \cdot \text{{weakClassifier}}_t(x)\right)\]
where: \\
$T$ represents the total number of weak classifiers (base estimators) in the AdaBoost ensemble\\
$\alpha_t$ is the weight or contribution of $\text{weakClassifier}_t$ to the final prediction, and \\ $\text{weakClassifier}_t(x)$ represents the prediction of the $t$-th weak classifier on input $x$.

\subsubsection{Model Training (GradientBoosting Classifier):}
 GradientBoosting Classifier starts with empty ensemble and initializes predictions to a constant value. Then, it iteratively fits new models to correct the errors made by the previous ensemble. Weak classifiers are trained using input features and the residuals between predicted and actual values. Pseudo-residuals, which represent the errors to be corrected, are calculated. Additional models are trained to predict these pseudo-residuals and are added to the ensemble. This process continues until a stopping criterion is met \cite{huang2018credit} \cite{castillo2019application} \cite{chauhan2020credit} \cite{wu2021credit}.
 
The equation for GradientBoostingClassifier in the context of RTDPA is different from XGB Classifier and CatBoost Classifier. In GradientBoostingClassifier, instead of using an explicit weighting scheme for weak learners, the predictions of weak learners are aggregated in a sequential manner. \\ The equation for $r$ in $R$, where $R$ represents the set of distinct row types in the dataset can be represented as:
\[\hat{y}_{\text{{new}}}(r) = \sum_{k=1}^{K_r} \alpha_k(r) \cdot f_k(r)(x_{\text{{new}}})\]
where,\\ $\alpha_k(r)$ represents the contribution of each weak learner $f_k(r)$ to the final prediction. The values of $\alpha_k(r)$ are determined during the training process, with each subsequent weak learner aiming to correct the errors made by the previous learners.

Therefore, the key difference lies in the way the weak learners are combined. While XGBClassifier and CatBoostClassifier use weighted ensembling, GradientBoostingClassifier employs additive ensembling where each weak learner contributes directly to the final prediction without explicit weights.

\section{Experimental Results}
In this section, we present the results of our analysis, which will serve as the basis for selecting the appropriate algorithm, the primary purpose of this work.
\subsection{Data}

\begin{wraptable}{r}{0.5\textwidth}
	\caption{Loan Classification}
	\label{tab:Loan Classification}
	\begin{tabular}{|l|l|l|}
		\hline
		\textbf{Loan Type} & \textbf{IRAC} & \textbf{Count} \\
		\hline
		\multirow{4}{*}{Agriculture Loan} & Standard & 17496 \\
		\cline{2-3}
		& Sub standard & 294 \\
		\cline{2-3}
		& Doubtful & 2577 \\
		\cline{2-3}
		& Loss & 210 \\
		\hline
		\multirow{4}{*}{Personal Loan} & Standard & 4398 \\
		\cline{2-3}
		& Sub standard & 126 \\
		\cline{2-3}
		& Doubtful & 129 \\
		\cline{2-3}
		& Loss & 3 \\
		\hline
	\end{tabular}
\end{wraptable}

 Our dataset consists of more than 25,000 samples from xxx Bank, with 81 attributes listed in Table\ref{tab:column_description}. The loan types in the dataset can be categorized into two distinct categories: Agriculture loans and Personal loans. Each loan entry is classified as either standard, substandard, doubtful, or loss as listed in table\ref{tab:Loan Classification}. However, for personal loans, there are only three samples classified as loss, so we have merged them into the doubtful category. To implement the Row-Type Dependent Predictive Analysis (RTDPA), we have separated the rows for each subcategory of loans and processed them individually.
 
\subsection{Data PreProcessing}

Features dryland and wetland are not applicable to personal loan, so we removed them from analysis of personal loan.  We also analysed missing values for each category and dropped features having missing values more then $70\%$ as listed in table\ref{tab:missing-values} below. \\ 

\begin{table}[H]	
	\caption{\% of Missing Values for Personal Loan and Agriculture Loan}
	\label{tab:missing-values}
	\begin{subtable}{0.4\textwidth}
		\centering
		\caption{Personal Loan}
		\begin{tabular}{|lcc|}
			\hline
			Variable   & Total Missing & \% Missing \\
			\hline
			OPINIONDT  & 4436          & 95.3 \\
			DIRFINFLG  & 4656          & 100.0 \\
			SANAUTCD   & 3838          & 82.4 \\
			DOCREVDT   & 4229          & 90.8 \\
			PRISECCD2  & 4406          & 94.6 \\
			RENEWALDT  & 4656          & 100.0 \\
			INSEXPDT   & 4439          & 95.3 \\
			TFRDT      & 4216          & 90.5 \\
			REASONCD   & 4181          & 89.8 \\
			RECALLDT   & 4216          & 90.5 \\
			WOSACD     & 4568          & 98.1 \\
			\hline
		\end{tabular}
	\end{subtable}%
	\begin{subtable}{0.7\textwidth}
		\centering
		\caption{Agriculture Loan}
		\begin{tabular}{|lcc|}
			\hline
			Variable   & Total Missing & \% Missing \\
			\hline
			OPINIONDT  & 19743         & 95.9 \\
			SANAUTCD   & 18067         & 87.8 \\
			DOCREVDT   & 19918         & 96.8 \\
			PRISECCD2  & 19668         & 95.6 \\
			UNIFUNFLG  & 15771         & 76.6 \\
			RENEWALDT  & 19956         & 97.0 \\
			INSEXPDT   & 20502         & 99.6 \\
			TFRDT      & 18090         & 87.9 \\
			REASONCD   & 18081         & 87.9 \\
			RECALLDT   & 18096         & 87.9 \\
			WOSACD     & 19579         & 95.1 \\
			\hline
		\end{tabular}
	\end{subtable}
\end{table}
\subsection{Feature Selection}

After preprocessing the data, we dropped features with unique values for both categories and replaced missing values, we addressed the challenge of high dimensionality by applying Principal Component Analysis (PCA). PCA is a statistical technique commonly used to explore interrelationships among variables and reduce dimensionality.

By projecting the data onto a new vector space defined by the eigenvectors of the original dataset, PCA allows us to assess variable importance while mitigating redundancies and high dimensionality. Through linear combinations of the original variables, PCA creates a new set of variables that reveal a clearer underlying structure within the data\cite{panigrahi2011}\cite{andras2016}.

To determine the optimal number of principal components that capture most of the data's variance, we plotted a scree plot. The scree plot helps identify the "elbow" or "knee" point, where the eigenvalues or variance explained by the components start to level off. This point indicates the number of components that should be retained since additional components provide diminishing returns in capturing additional variance\cite{huynh2017}. For our model, we identified 43 principal components for personal data and 38 principal components for agriculture data.

\hspace{4cm}
\begin{figure}[H]
	\centering
	\begin{subfigure}{0.5\textwidth}
		\centering
		\includegraphics[width=\linewidth]{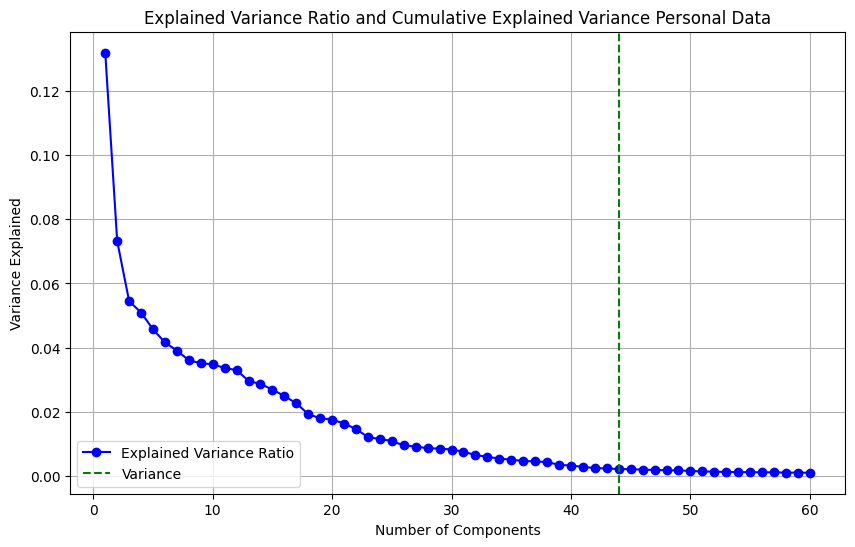} 
		\caption{PCA Personal Data}
		\label{fig:subfig1}
	\end{subfigure}%
	\begin{subfigure}{0.5\textwidth}
		\centering
		\includegraphics[width=\linewidth]{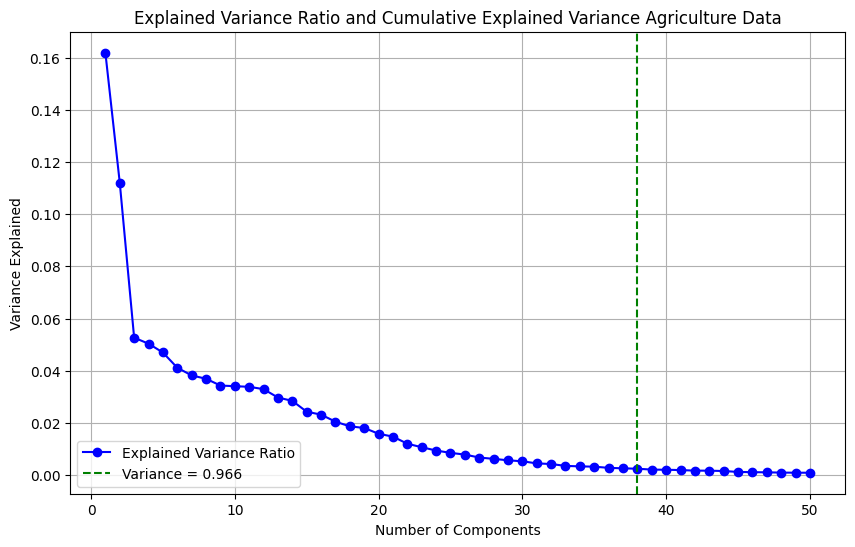} 
		\caption{PCA Agriculture Data}
		\label{fig:subfig2}
	\end{subfigure}
	\caption{PCA}
	\label{fig:mainfig}
\end{figure}

\subsection{Data Augmentation}

Given the highly imbalanced nature of our data, we conducted experiments involving various transformations and modifications to our existing dataset. In order to address the class imbalance issue, we utilized the Synthetic Minority Over-sampling Technique (SMOTE) and its variants \cite{chawla2002smote}.

SMOTE (Synthetic Minority Over-sampling Technique): SMOTE generates synthetic samples for the minority class by interpolating between existing minority class instances. It creates new synthetic instances along the line segments connecting nearest neighbors of a minority class instance \cite{chawla2002smote}.

ADASYN (Adaptive Synthetic Sampling): ADASYN is an extension of SMOTE that takes into account the density distribution of the minority class. It generates more synthetic samples for minority class instances that are harder to learn, i.e., instances located in regions with a lower density of the minority class \cite{he2008adasyn}.

SMOTETomek: SMOTETomek combines the SMOTE oversampling technique with the Tomek links undersampling technique. Tomek links are pairs of instances from different classes that are nearest neighbors of each other. SMOTETomek first applies SMOTE to oversample the minority class and then removes the Tomek links to remove overlapping instances\cite{batista2006balancing}.

SMOTEENN (SMOTE Edited Nearest Neighbors): SMOTEENN is another combination of SMOTE oversampling and an undersampling technique called Edited Nearest Neighbors (ENN). SMOTEENN first applies SMOTE to generate synthetic samples and then uses ENN to remove instances that are misclassified by the nearest neighbors classifier\cite{batista2004study}.

These techniques aim to address the class imbalance problem in datasets by either oversampling the minority class (SMOTE and ADASYN) or combining oversampling and undersampling approaches (SMOTETomek and SMOTEENN). They help to create a more balanced and representative dataset for training machine learning models on imbalanced datasets

\subsection{Evaluation Matrix}

TP, FP, FN, and TN are commonly used in the context of binary classification. However, they can be extended and adapted for multiclass classification as well.

In multiclass classification, we have more than two classes, and each class can be considered as either positive or negative in relation to the other classes. The terms TP, FP, FN, and TN can be defined for each class pair.

Here's how the terms can be defined for a specific class (let's say Class A) in a multiclass classification scenario:\\
TP (True Positives): The instances that belong to Class A and are correctly predicted as Class A.\\
FP (False Positives): The instances that do not belong to Class A but are incorrectly predicted as Class A.\\
FN (False Negatives): The instances that belong to Class A but are incorrectly predicted as not Class A.\\
TN (True Negatives): The instances that do not belong to Class A and are correctly predicted as not Class A.\\
Similarly, these terms can be defined for each class in a multiclass classification problem.

In summary, while TP, FP, FN, and TN are commonly associated with binary classification, they can be adapted for multiclass classification by considering the performance of the model for each class pair.
Accuracy: Accuracy is a commonly used evaluation metric in classification tasks\cite{sokolova2009systematic}\cite{zhang2012evaluating}. It measures the overall correctness of the model's predictions by calculating the ratio of correct predictions to the total number of predictions. It is calculated as:
\[\text{Accuracy} = \frac{\text{Number of Correct Predictions}}{\text{Total Number of Predictions}}\]

Accuracy provides an indication of the model's ability to classify instances correctly across all classes. However, it may not be suitable when dealing with imbalanced datasets, where the number of instances in different classes varies significantly.

Precision: Precision is the ratio of true positive predictions to the total number of positive predictions\cite{pleiss2017precision}. It measures the accuracy of positive predictions and is calculated as:
\[\text{Precision} = \frac{\text{True Positives}}{\text{True Positives} + \text{False Positives}}\]

Recall: Recall, also known as sensitivity or true positive rate, is the ratio of true positive predictions to the total number of actual positive instances in the data. It measures the ability to identify positive instances correctly and is calculated as:

\[\text{Recall} = \frac{\text{True Positives}}{\text{True Positives} + \text{False Negatives}}\]

F1 Score: The F1 score is the harmonic mean of precision and recall. It provides a balanced measure of the model's performance by considering both precision and recall. It is calculated as:

\[\text{F1 Score} = 2 \times \frac{\text{Precision} \times \text{Recall}}{\text{Precision} + \text{Recall}}\]

ROC AUC: ROC AUC (Receiver Operating Characteristic Area Under the Curve) is a measure of the model's ability to discriminate between positive and negative classes. It is calculated by plotting the true positive rate against the false positive rate at various classification thresholds and calculating the area under the curve\cite{voorhees2001evaluation}\cite{sokolova2009systematic}.

Kappa: Cohen's kappa coefficient is a statistical measure of inter-rater agreement or reliability for categorical variables\cite{cohen1960}. In the context of classification models, it quantifies the agreement between predicted and true class labels, considering the possibility of agreement occurring by chance. It ranges from -1 to 1, where a value closer to 1 indicates higher agreement than expected by chance.

These evaluation metrics are commonly used to assess the performance of classification models and provide insights into their accuracy, sensitivity, and overall effectiveness.

\subsection{Decision Tree Results}

\begin{figure}[H]
	\centering
	\includegraphics[height=0.7\textwidth,width=0.8\textwidth]{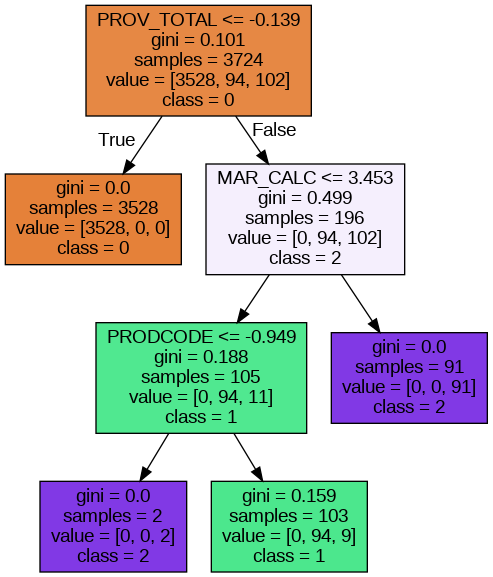}
	\caption{Decision Tree Personal Data}
	\label{fig:DT_Personal}
\end{figure}

\begin{figure}[H]
	\includegraphics[angle=90,height=22cm,width=1.1\textwidth]{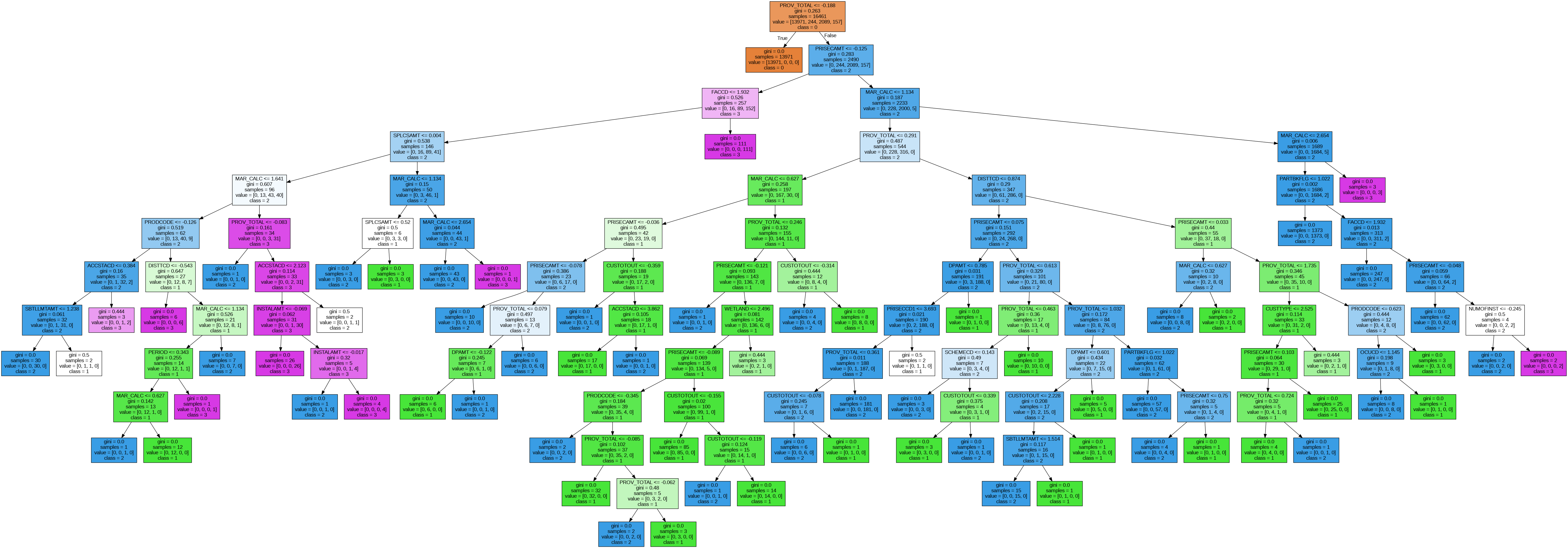}
	\caption{Decision Tree Personal Data}
	\label{fig:DT_Agri}
\end{figure}

Our algorithm, RTDPA, creates two distinct decision trees, Figure \ref{fig:DT_Personal} and Figure \ref{fig:DT_Agri}, for each row type $r$ in $R$ in the dataset $D$. The analysis reveals that the decision tree associated with personal credit rows requires fewer attributes compared to the agriculture decision tree.

\hspace{5cm}

The tables \ref{tab:DT model_performance for Personal Loan} and \ref{tab:DT model_performance for Agriculture Loan} presents the performance of different classifiers for the personal loan and agriculture loan dataset. The metrics evaluated include train accuracy, test accuracy, precision, recall, F1 score, ROC AUC, Cohen's kappa, and running time. \\Here are some summary findings:\\
\\
For personal risk, RandomForest Classifier achieves the highest train accuracy of 1.0, indicating a perfect fit to the training data.
XGB Classifier achieves the highest test accuracy of 0.9989, indicating strong generalization performance.
XGB Classifier also performs best in terms of precision, recall, F1 score, ROC AUC, and Cohen's kappa, indicating its overall superiority in classification performance. DecisionTree Classifier has the shortest running time of 0.55 seconds.\\
Based on these results, XGB Classifier emerges as the top-performing classifier for the personal loan dataset, demonstrating high accuracy, precision, recall, and F1 score. It also achieves a perfect ROC AUC and a high Cohen's kappa, suggesting robust and reliable classification results.\\

\begin{table}[H]
	\centering
	\caption{Model Performance for Personal Loan}
	\hspace{4cm}
	\label{tab:DT model_performance for Personal Loan}
	\begin{tabular}{|p{2.6cm}|p{1.5cm}|p{1.5cm}|p{1.5cm}|p{1cm}|p{1cm}|p{1cm}|p{1.3cm}|p{1.7cm}|}
		\hline
		\textbf{Classifier} & \textbf{Train} & \textbf{Test} & \textbf{Precision} & \textbf{Recall} & \textbf{F1-} & \textbf{ROC} & \textbf{Cohen's} & \textbf{Running} \\
		& \textbf{Accuracy} & \textbf{Accuracy} & & & \textbf{Score} & \textbf{AUC} & \textbf{Kappa} & \textbf{Time} \\
		\hline
			DecisionTree & 0.9976 & 0.9968 & 0.9683 & 0.9674 & 0.9677 & 0.9938 & 0.9745 & 00:00.5796 \\
			RandomForest & 1 & 0.9968 & 0.9697 & 0.9688 & 0.9677 & 1 & 0.9745 &  01:08.9367 \\
			ExtraTrees & 1 & 0.9914 & 0.9374 & 0.9160 & 0.9224 & 0.9992 & 0.9310 & 00:01.3297 \\
			LGBM & 1 & 0.9979 & 0.9792 & 0.9792 & 0.9785 & 0.9999 & 0.9830 &  00:01.6655 \\
			XGB & 1 & 0.9989 & 0.9892 & 0.9896 & 0.9892 & 1 & 0.9915 & 00:05.9198 \\
			CatBoost & 1 & 0.9979 & 0.9804 & 0.9778 & 0.9784 & 1 & 0.9830 &  00:13.2509 \\
			GradientBoosting & 1 & 0.9957 & 0.9569 & 0.9569 & 0.9569 & 0.9999 & 0.9660 &  00:56.4960 \\
			AdaBoost & 0.9970 & 0.9979 & 0.9804 & 0.9778 & 0.9784 & 0.9992 & 0.9830 & 01:43.1385 \\			
		\hline
	\end{tabular}
	\end{table}

Best estimator based on Train Accuracy:RandomForestClassifier (Train Accuracy: 1.0)\\
Best estimator based on Test Accuracy: XGBClassifier (Test Accuracy: 0.9989270386266095)\\
Best estimator based on Precision: XGBClassifier (Precision: 0.989247311827957)\\
Best estimator based on Recall: XGBClassifier (Recall: 0.9895833333333334)\\
Best estimator based on F1 Score: XGBClassifier (F1 Score: 0.9892445138346778)\\
Best estimator based on ROC AUC: XGBClassifier (ROC AUC: 1.0)\\
Best estimator based on Cohen's Kappa: XGBClassifier (Cohen's Kappa: 0.9915119943170434)\\
Best estimator based on Running Time: DecisionTreeClassifier (Running Time: 0 days 00:00:00.550333)\\
\\
For Agriculture loan risk analysis, RandomForestClassifier achieved the highest Train Accuracy of 1.0, indicating a perfect fit to the training data.  LGBMClassifier achieved the highest Test Accuracy of 0.9966, suggesting strong generalization performance on unseen data.  ExtraTreesClassifier achieved the highest Precision of 0.9725, indicating a high proportion of correct positive predictions.  LGBMClassifier achieved the highest Recall of 0.9578, indicating a high proportion of correctly identified positive instances.  LGBMClassifier achieved the highest F1 Score of 0.9618, which is a balanced measure of Precision and Recall.  LGBMClassifier achieved the highest ROC AUC of 0.9996, indicating excellent discrimination between positive and negative instances.
LGBMClassifier achieved the highest Cohen's Kappa of 0.9865, which measures agreement beyond chance between predicted and actual labels.  DecisionTreeClassifier had the shortest Running Time of 2.26 seconds.\\
Based on the analysis, LGBMClassifier emerges as the best performer for the Agriculture Loan dataset, with high accuracy, precision, recall, and F1 Score, as well as strong discrimination power and agreement with the actual labels. However, other classifiers, such as ExtraTreesClassifier and XGBClassifier, also show competitive performance and can be considered as alternative options.\\

\begin{table}[h]
	\centering
	\caption{Model Performance for Agriculture Loan}
	\hspace{4cm}
	\label{tab:DT model_performance for Agriculture Loan}
	\begin{tabular}{|p{2.6cm}|p{1.5cm}|p{1.5cm}|p{1.5cm}|p{1cm}|p{1cm}|p{1cm}|p{1.3cm}|p{1.7cm}|}
		\hline
		\textbf{Classifier} & \textbf{Train} & \textbf{Test} & \textbf{Precision} & \textbf{Recall} & \textbf{F1-} & \textbf{ROC} & \textbf{Cohen's} & \textbf{Running} \\
		& \textbf{Accuracy} & \textbf{Accuracy} & & & \textbf{Score} & \textbf{AUC} & \textbf{Kappa} & \textbf{Time} \\
		\hline
		DecisionTree & 0.9996 & 0.9959 & 0.9521 & 0.9520 & 0.9520 & 0.9761 & 0.9836 &  00:02.26 \\
		LGBM & 1 & 0.9966 & 0.9660 & 0.9578 & 0.9618 & 0.9996 & 0.9865 & 00:06.13 \\
		ExtraTrees & 1 & 0.9903 & 0.9725 & 0.8387 & 0.8914 & 0.9988 & 0.9608 & 00:07.15 \\
		CatBoost & 0.9980 & 0.9947 & 0.9596 & 0.9228 & 0.9402 & 0.9991 & 0.9788 & 00:40.66 \\
		XGB & 1 & 0.9964 & 0.9652 & 0.9531 & 0.9590 & 0.9993 & 0.9855 & 00:48.45 \\
		RandomForest & 1 & 0.9939 & 0.9608 & 0.9036 & 0.9296 & 0.9993 & 0.9759 & 03:28.33 \\
		AdaBoost & 0.9844 & 0.9810 & 0.9083 & 0.6789 & 0.6679 & 0.9967 & 0.9244 &  06:12.49 \\
		GradientBoosting & 0.9995 & 0.9942 & 0.9443 & 0.9269 & 0.9347 & 0.9994 & 0.9769 & 07:48.37 \\
	
		\hline
	\end{tabular}
\end{table}

Best estimator based on Train Accuracy: RandomForestClassifier (Train Accuracy: 1.0)\\
Best estimator based on Test Accuracy: LGBMClassifier (Test Accuracy: 0.9965986394557823)\\
Best estimator based on Precision: ExtraTreesClassifier (Precision: 0.9724740018034564)\\
Best estimator based on Recall: LGBMClassifier (Recall: 0.9577752861119704)\\
Best estimator based on F1 Score: LGBMClassifier (F1 Score: 0.961805883278276)\\
Best estimator based on ROC AUC: LGBMClassifier (ROC AUC: 0.9995529854882718)\\
Best estimator based on Cohen's Kappa: LGBMClassifier (Cohen's Kappa: 0.986509682294871)\\
Best estimator based on Running Time: DecisionTreeClassifier (Running Time: 0 days 00:00:02.262857)\\

\subsection{Logistic Regression, Naive Bayes, Neural Network, K-NN, Random Forest and SVM Results}

we compared the performance of different classifiers including Logistic Regression, Gaussian NB, Neural Network, K-Nearest Neighbors, and Random Forest and draw inferences for both personal and agriculture loan data Table \ref{tab:LR_model_performance}.\\\\
For "Personal" risk analysis,  Logistic Regression (SMOTE) and Logistic Regression (SMOTETomek) consistently exhibit high performance across all metrics. They achieve high accuracy, precision, recall, F1-score, ROC AUC, and Cohen's Kappa values. These models are well-suited for predicting personal category-related data.\\
Neural Network models also perform exceptionally well for the "Personal" category, achieving near-perfect accuracy. They demonstrate high precision, recall, F1-score, ROC AUC, and Cohen's Kappa. However, Neural Network models have longer running times compared to other models.\\
Random Forest models also show strong performance in terms of accuracy, precision, recall, F1-score, and ROC AUC. These models can be considered as an alternative to Logistic Regression models, with reasonable running times.\\
K-Nearest Neighbors models achieve moderate performance for the "Personal" category, with relatively shorter running times. Although not as high as Logistic Regression, Neural Network, or Random Forest models, they still provide reasonable results.\\
Gaussian NB models have the shortest running times among all the models, but they demonstrate lower performance metrics for the "Personal" category. Their accuracy, precision, recall, F1-score, and ROC AUC values are comparatively lower than other models.\\\\
For the "Agriculture" risk analysis,\\
Similar to the "Personal" category, Logistic Regression (SMOTE) and Logistic Regression (SMOTETomek) models consistently show high performance across all metrics. They achieve high accuracy, precision, recall, F1-score, ROC AUC, and Cohen's Kappa values. These models are reliable for predicting agriculture-related data.

Neural Network models perform exceptionally well for the "Agriculture" category, with high accuracy, precision, recall, F1-score, ROC AUC, and Cohen's Kappa values. However, as mentioned before, they have longer running times compared to other models.

Random Forest models also exhibit strong performance for the "Agriculture" category, with high accuracy, precision, recall, F1-score, and ROC AUC. They offer reasonable running times, making them a viable choice for this category.

K-Nearest Neighbors models achieve moderate performance for the "Agriculture" category, similar to their performance in the "Personal" category. They provide decent results with shorter running times.

Gaussian NB models have the shortest running times, but they show lower performance metrics for the "Agriculture" category. Their accuracy, precision, recall, F1-score, and ROC AUC values are comparatively lower than other models.

In summary, Logistic Regression models (specifically with SMOTE and SMOTETomek techniques) consistently perform well for both the "Personal" and "Agriculture" categories. Neural Network models achieve high accuracy in both categories but have longer running times. Random Forest models offer a good balance between performance and running time. K-Nearest Neighbors models provide moderate performance with shorter running times. Gaussian NB models have the shortest running times but sacrifice some performance.

\begin{sidewaystable}[htbp]
	\centering
	\caption{Model Performance for Logistic Regression, Gaussian NB, Neural Network, KNN, Random Forest}
	\label{tab:LR_model_performance}
	\begin{tabular}{|p{3.1cm}|p{1.6cm}|p{1.7cm}|p{1.5cm}|p{1.5cm}|p{1cm}|p{1cm}|p{1.0cm}|p{1.3cm}|p{1.7cm}|}
		\hline
\textbf{Classifier} & \textbf{Row Type} & \textbf{Train Accuracy} & \textbf{Test Accuracy} & \textbf{Precision} & \textbf{Recall} & \textbf{F1 Score} & \textbf{ROC AUC} & \textbf{Cohen's Kappa} & \textbf{Running Time} \\
\hline
	\multirow{2}{*}{\shortstack[l] {Logistic Regression \\ (SMOTE)}} &Personal & 0.9648 & 0.9442 & 0.9647 & 0.9442 & 0.9518 & 0.9714 & 0.6559 & 00.703323 \\
	& Agriculture & 0.9648 & 0.9442 & 0.9647 & 0.9442 & 0.9518 & 0.9714 & 0.6559 & 00.703323 \\
	\hline
	\multirow{2}{*}{\shortstack[l]{Logistic Regression \\(ADASYN)}}& Personal & 0.9541 & 0.9313 & 0.9587 & 0.9313 & 0.9416 & 0.9655 & 0.5957 & 00.817455 \\
	& Agriculture  & 0.9541 & 0.9313 & 0.9587 & 0.9313 & 0.9416 & 0.9655 & 0.5957 & 00.817455 \\
	\hline
	\multirow{2}{*}{\shortstack[l]{Logistic Regression\\ (SMOTETomek)}} & Personal& 0.9683 & 0.9485 & 0.9660 & 0.9485 & 0.9551 & 0.9727 & 0.6726 & 00.819621 \\
	& Agriculture & 0.9683 & 0.9485 & 0.9660 & 0.9485 & 0.9551 & 0.9727 & 0.6726 & 00.819621 \\
	\hline
	\multirow{2}{*}{\shortstack[l]{Logistic Regression\\ (SMOTEENN)}}& Personal & 0.9568 & 0.9388 & 0.9636 & 0.9388 & 0.9480 & 0.9683 & 0.6337 & 01.304687 \\
	& Agriculture & 0.9568 & 0.9388 & 0.9636 & 0.9388 & 0.9480 & 0.9683 & 0.6337 & 00.703323 \\
	\hline
	\multirow{2}{*}{\shortstack[l]{GaussianNB \\ (SMOTE)}}& Personal & 0.9296 & 0.9195 & 0.9626 & 0.9195 & 0.9364 & 0.9556 & 0.5564 & 00.018716 \\
	& Agriculture & 0.5553 & 0.5428 & 0.9431 & 0.5428 & 0.6802 & 0.8845 & 0.2111 & 00.018716 \\
	\hline
	\multirow{2}{*}{\shortstack[l]{GaussianNB\\ (ADASYN)}} & Personal& 0.9141 & 0.8970 & 0.9679 & 0.8970 & 0.9208 & 0.9662 & 0.4740 & 00.014647 \\
	& Agriculture & 0.4799 & 0.4674 & 0.8923 & 0.4674 & 0.6018 & 0.8220 & 0.1339 & 00.014647 \\
	\hline
	\multirow{2}{*}{\shortstack[l]{ GaussianNB \\(SMOTETomek)}} & Personal & 0.9305 & 0.9163 & 0.9632 & 0.9163 & 0.9345 & 0.9568 & 0.5483 & 00.007762 \\
	& Agriculture & 0.5724 & 0.5590 & 0.9487 & 0.5590 & 0.6942 & 0.8883 & 0.2330 & 00.007762 \\
	\hline
	\multirow{2}{*}{\shortstack[l]{GaussianNB\\ (SMOTEENN)}} & Personal & 0.9476 & 0.9378 & 0.9650 & 0.9378 & 0.9483 & 0.9582 & 0.6228 & 00.007857 \\
	& Agriculture & 0.6060 & 0.5931 & 0.9537 & 0.5931 & 0.7228 & 0.9086 & 0.2571 & 00.007857 \\
	\hline
	\multirow{2}{*}{\shortstack[l]{Neural Network\\ (SMOTE)}} & Personal & 0.9997 & 0.9882 & 0.9879 & 0.9882 & 0.9878 & 0.9779 & 0.9044 & 10.906960 \\
	& Agriculture  & 0.9956 & 0.9740 & 0.9759 & 0.9740 & 0.9749 & 0.9728 & 0.8985 & 10.906960 \\
	\hline
	\multirow{2}{*}{\shortstack[l]{Neural Network\\ (ADASYN)}}& Personal & 0.9997 & 0.9861 & 0.9858 & 0.9861 & 0.9858 & 0.9731 & 0.8888 & 12.241614 \\
	& Agriculture & 0.9960 & 0.9725 & 0.9733 & 0.9725 & 0.9728 & 0.9611 & 0.8930 & 12.241614 \\
	\hline
	\multirow{2}{*}{\shortstack[l]{Neural Network \\(SMOTETomek)}}& Personal & 0.9995 & 0.9882 & 0.9886 & 0.9882 & 0.9879 & 0.9768 & 0.9059 & 06.725499 \\
	& Agriculture & 0.9949 & 0.9735 & 0.9747 & 0.9735 & 0.9741 & 0.9649 & 0.8967 & 06.725499 \\
	\hline
	\multirow{2}{*}{\shortstack[l]{Neural Network \\(SMOTEENN)}} &Personal & 0.9914 & 0.9763 & 0.9788 & 0.9763 & 0.9772 & 0.9674 & 0.8252 & 11.619167\\
	&Agriculture & 0.9834 & 0.9625 & 0.9700 & 0.9625  & 0.9656  & 0.9672 & 0.8590 & 11.619167\\	   
	\hline
	\multirow{2}{*}{\shortstack[l]{K-Nearest Neighbors\\ (SMOTE)}} & Personal &0.9744 &	0.9474 &	0.9602 & 0.9474 & 0.9526 &	0.8956 & 0.6485  & 00.001439 \\
	& Agriculture & 0.9664	& 0.9312 & 0.95033	&0.9312	&0.9395	&0.8802 & 0.7542 & 00.001439 \\
	\hline
	\multirow{2}{*}{\shortstack[l]{K-Nearest Neighbors\\ (ADASYN)}} & Personal & 0.9736 & 0.9452 &	0.9609	& 0.9452 & 0.9515 & 0.8838 & 0.6433 & 00.004408 \\
	& Agriculture &	0.9580 & 0.9198 & 0.9429 & 0.9198 & 0.9293 & 0.8733 & 0.7220 & 00.004408 \\
	\hline
	\multirow{2}{*}{\shortstack[l]{K-Nearest Neighbors \\(SMOTETomek)}} & Personal & 0.9739 & 0.9485 & 0.9602 & 0.9485 & 0.9533 & 0.8889 & 0.6512 & 00.001705 \\
	& Agriculture & 0.9647 & 0.9310 & 0.9504 & 0.9310 & 0.9394 & 0.8807 & 0.7543  & 00.001705 \\
	\hline
	\multirow{2}{*}{\shortstack[l]{K-Nearest Neighbors \\(SMOTEENN)}} & Personal  &0.9608 & 0.9345 & 0.9577 & 0.9345 & 0.9436 & 0.8788 & 0.5987 &00.001644 \\
	& Agriculture & 0.9470 & 0.9176 & 0.9482 & 0.9176 & 0.9306 & 0.8775 & 0.7193  & 00.001644 \\
	\hline
	\multirow{2}{*}{\shortstack[l]{Random Forest\\ (SMOTE)}} & Personal & 1.0000 & 0.9893 & 0.9893 & 0.9893 & 0.9889 & 0.9739 & 0.9138 & 06.744184 \\
	& Agriculture & 1.0000 & 0.9742 & 0.9750 & 0.9742 & 0.9745 & 0.9761 & 0.8965 & 06.744184 \\
	\hline
	\multirow{2}{*}{\shortstack[l]{Random Forest\\ (ADASYN)}} & Personal & 1.0000 & 0.9870 & 0.9870 & 0.9871 & 0.9868 & 0.9780 &0.888 & 06.604297 \\
	& Agriculture & 1.0000 & 0.9718 & 0.9728 & 0.9718 & 0.9722 & 0.9716 & 0.8973 & 06.604297 \\
	\hline
	\multirow{2}{*}{\shortstack[l]{Random Forest\\ (SMOTETomek)}} & Personal & 0.9997 & 0.9849 & 0.9849 & 0.9849 & 0.9848 & 0.9815 &0.8820 & 05.968728 \\
	& Agriculture & 0.9991 & 0.9740 & 0.9750 & 0.9740 & 0.9744 & 0.9750 &0.8955	 & 05.968728 \\
	\hline
	\multirow{2}{*}{\shortstack[l]{Random Forest\\ (SMOTEENN)}} & Personal & 0.9903 & 0.9785 & 0.9809 & 0.9785 & 0.9794 & 0.9778 & 0.8422& 06.091716 \\
	& Agriculture & 0.9837 & 0.9662 & 0.9722 & 0.9662 & 0.9688 & 0.9740& 0.8697 & 06.091716 \\
	\hline
	\end{tabular}
\end{sidewaystable}

\begin{sidewaystable}[htpb]
	\centering
	\caption{Model Performance for SVM }
	\label{tab:SVM_model_performance}
	\begin{tabular}{|p{4.9cm}|p{1.8cm}|p{1.7cm}|p{1.5cm}|p{1.5cm}|p{1cm}|p{1cm}|p{1.3cm}|p{1.7cm}|p{1.7cm}|}
		\hline
		\textbf{Classifier} & \textbf{Row Type} & \textbf{Train Accuracy} & \textbf{Test Accuracy} & \textbf{Precision} & \textbf{Recall} & \textbf{F1 Score} & \textbf{ROC AUC} & \textbf{Cohen's Kappa} & \textbf{Running Time} \\
		\hline

\multirow{2}{*}{SVM (SMOTE, Linear)} & Personal & 0.9820 & 0.9646 & 0.9716 & 0.9646 & 0.9673 & 0.9617 & 0.7521 & 11.3272 \\
& Agriculture & 0.9145 & 0.8858 & 0.9489 & 0.8858 & 0.9108 & 0.9651 & 0.6514 & 11.3272 \\
\hline
\multirow{2}{*}{\shortstack[l]{SVM (ADASYN, Linear)}} & Personal & 0.9761 & 0.9592 & 0.9690 & 0.9592 & 0.9629 & 0.9366 & 0.7239 & 12.1966 \\
& Agriculture & 0.8764 & 0.8610 & 0.9411 & 0.8610 & 0.8912 & 0.9575 & 0.6017 & 12.1966 \\
\hline
\multirow{2}{*}{\shortstack[l]{SVM (SMOTETomek, Linear)}} & Personal & 0.9823 & 0.9657 & 0.9725 & 0.9657 & 0.9683 & 0.9547 & 0.7596 & 08.5337 \\
& Agriculture & 0.9159 & 0.8836 & 0.9502 & 0.8836 & 0.9104 & 0.9637 & 0.6465 & 08.5337 \\
\hline
\multirow{2}{*}{\shortstack[l]{SVM (SMOTEENN, Linear)}} & Personal & 0.9791 & 0.9582 & 0.9690 & 0.9582 & 0.9624 & 0.9665 & 0.7166 & 08.1002 \\
& Agriculture & 0.9071 & 0.8861 & 0.9577 & 0.8861 & 0.9157 & 0.9676 & 0.6536 & 08.1002 \\
\hline
\multirow{2}{*}{\shortstack[l]{SVM (SMOTE, Poly)}} & Personal & 0.9944 & 0.9807 & 0.9812 & 0.9807 & 0.9803 & 0.9663 & 0.8495 & 06.6656 \\
& Agriculture & 0.9452 & 0.9264 & 0.9586 & 0.9264 & 0.9401 & 0.9498 & 0.7456 & 06.6656 \\
\hline
\multirow{2}{*}{\shortstack[l]{SVM (ADASYN, Poly)}} & Personal & 0.9815 & 0.9732 & 0.9770 & 0.9732 & 0.9746 & 0.9711 & 0.8056 & 12.7388 \\
& Agriculture & 0.8798 & 0.8756 & 0.9346 & 0.8756 & 0.8967 & 0.9503 & 0.6259 & 12.7388 \\
\hline
\multirow{2}{*}{\shortstack[l]{SVM (SMOTETomek, Poly)}} & Personal & 0.9944 & 0.9839 & 0.9839 & 0.9839 & 0.9836 & 0.9699 & 0.8737 & 08.4919 \\
& Agriculture & 0.9450 & 0.9281 & 0.9606 & 0.9281 & 0.9421 & 0.9486 & 0.7493 & 08.4919 \\
\hline
\multirow{2}{*}{\shortstack[l]{SVM (SMOTEENN, Poly)}} & Personal & 0.9863 & 0.9742 & 0.9769 & 0.9742 & 0.9749 & 0.9614 & 0.8108 & 08.0531 \\
& Agriculture & 0.9317 & 0.9082 & 0.9545 & 0.9082 & 0.9277 & 0.9540 & 0.6994 & 08.0531 \\
\hline
\multirow{2}{*}{\shortstack[l]{SVM (SMOTE, RBF)}} & Personal & 0.9914 & 0.9882 & 0.9883 & 0.9882 & 0.9877 & 0.9776 & 0.9037 & 05.2920 \\
& Agriculture & 0.9667 & 0.9524 & 0.9661 & 0.9524 & 0.9581 & 0.9723 & 0.8235 & 05.2920 \\
\hline
\multirow{2}{*}{\shortstack[l]{SVM (ADASYN, RBF)}} & Personal & 0.9890 & 0.9828 & 0.9825 & 0.9828 & 0.9826 & 0.9772 & 0.8621 & 17.6273 \\
& Agriculture & 0.9367 & 0.9308 & 0.9527 & 0.9308 & 0.9390 & 0.9675 & 0.7597 & 17.6273 \\
\hline
\multirow{2}{*}{\shortstack[l]{SVM (SMOTETomek, RBF)}} & Personal & 0.9914 & 0.9871 & 0.9878 & 0.9871 & 0.9868 & 0.9791 & 0.8966 & 04.5377 \\
& Agriculture & 0.9659 & 0.9519 & 0.9656 & 0.9519 & 0.9576 & 0.9734 & 0.8219 & 04.5377 \\
\hline
\multirow{2}{*}{\shortstack[l]{SVM (SMOTEENN, RBF)}} & Personal & 0.9842 & 0.9807 & 0.9825 & 0.9807 & 0.9809 & 0.9797 & 0.8518 & 06.1589 \\
& Agriculture & 0.9551 & 0.9458 & 0.9639 & 0.9458 & 0.9533 & 0.9751 & 0.8040 & 06.1589 \\
\hline
\multirow{2}{*}{\shortstack[l]{SVM (SMOTE, Sigmoid)}} & Personal & 0.9398 & 0.9185 & 0.9609 & 0.9185 & 0.9349 & 0.9678 & 0.5576 & 10.3033 \\
& Agriculture & 0.7754 & 0.7911 & 0.9326 & 0.7911 & 0.8491 & 0.9077 & 0.4658 & 10.3033 \\
\hline
\multirow{2}{*}{\shortstack[l]{SVM (ADASYN, Sigmoid)}} & Personal & 0.9184 & 0.8981 & 0.9531 & 0.8981 & 0.9188 & 0.9538 & 0.4898 & 16.3337 \\
& Agriculture & 0.7261 & 0.7264 & 0.9038 & 0.7264 & 0.7951 & 0.8994 & 0.3537 & 16.3337 \\
\hline
\multirow{2}{*}{\shortstack[l]{SVM (ADASYN, Sigmoid)}} & Personal & 0.9184 & 0.8981 & 0.9531 & 0.8981 & 0.9188 & 0.9538 & 0.4898 & 16.3337 \\
& Agriculture & 0.7261 & 0.7264 & 0.9038 & 0.7264 & 0.7951 & 0.8994 & 0.3537 & 16.3337 \\
\hline
\multirow{2}{*}{\shortstack[l]{SVM (SMOTETomek, Sigmoid)}} & Personal & 0.9398 & 0.9199 & 0.9610 & 0.9199 & 0.9361 & 0.9689 & 0.5604 & 09.0145 \\
& Agriculture & 0.7756 & 0.7928 & 0.9324 & 0.7928 & 0.8492 & 0.9076 & 0.4685 & 09.0145 \\
\hline
\multirow{2}{*}{\shortstack[l]{SVM (SMOTEENN, Sigmoid)}} & Personal & 0.9421 & 0.9239 & 0.9634 & 0.9239 & 0.9392 & 0.9706 & 0.5708 & 08.5702 \\
& Agriculture & 0.7914 & 0.8044 & 0.9337 & 0.8044 & 0.8693 & 0.9107 & 0.5064 & 08.5702 \\
\hline
	\end{tabular}
\end{sidewaystable}

\subsection{SVM Results}

The table \ref{tab:SVM_model_performance} presents the performance of various SVM models with different kernels (linear, poly, RBF, sigmoid) and sampling techniques (SMOTE, ADASYN, SMOTETomek, SMOTEENN) for both personal and agriculture datasets. \\\\
For the SVM models with a linear kernel, the performance is generally good across all sampling techniques. In the personal dataset, the models achieve high train and test accuracies, ranging from 0.9761 to 0.9823 and 0.9582 to 0.9657, respectively. The precision scores range from 0.9690 to 0.9725, indicating a high level of correctly classified positive instances. The recall scores are consistently high, ranging from 0.9592 to 0.9657, reflecting the models' ability to correctly identify positive instances. The F1 scores range from 0.9629 to 0.9683, which indicates a balanced trade-off between precision and recall.\\\\
Similarly, in the agriculture dataset, the SVM models with a linear kernel exhibit good performance. The train accuracies range from 0.8764 to 0.9159, and the test accuracies range from 0.8610 to 0.8836. The precision scores range from 0.9411 to 0.9502, indicating a high level of correctly classified positive instances. The recall scores range from 0.8610 to 0.8836, reflecting the models' ability to correctly identify positive instances. The F1 scores range from 0.8912 to 0.9104, indicating a balanced performance in terms of precision and recall.\\\\
The SVM models with polynomial, RBF, and sigmoid kernels also demonstrate promising performance, with varying degrees of accuracy, precision, recall, and F1 scores. Overall, the SVM models exhibit strong performance in classifying both personal and agriculture datasets, and the choice of kernel and sampling technique can be tailored based on the specific requirements.\\\\

\section{Conclusion}
	
Our adaptive modeling approach for row-type dependent predictive analysis offers a systematic and tailored methodology to address the heterogeneity of rows within a dataset. By recognizing and accommodating distinct row types, our approach enhances the accuracy and effectiveness of predictive models, enabling more insightful and accurate predictions across diverse datasets.     
\bibliographystyle{unsrt}
\bibliography{BankML}

		\begin{longtable}{|c|l|p{8cm}|}
			\caption{Column Description} \label{tab:column_description} \\
			\hline
			\textbf{COLUMN NO} & \textbf{COLUMN NAME} & \textbf{DESCRIPTION} \\
			\hline
			\endfirsthead
			
			\multicolumn{3}{c}%
			{{\tablename\ \thetable{} -- continued from previous page..}} \\
		
			\hline	
			\textbf{COLUMN NO} & \textbf{COLUMN NAME} & \textbf{DESCRIPTION} \\
			\hline
			\endhead
			
			\hline \multicolumn{3}{r}{{Continued on next page}} \\ \hline
			\endfoot
			
			\hline
			\endlastfoot
		
		1 & Q & Quarter \\
		2 & BRCD & Branch Code \\
		3 & CUSTID & Customer ID \\
		4 & ACCTID & Account ID \\
		5 & SEGCD & Segment Code \\
		6 & ORGCD & Org. Code \\
		7 & STFCD & Staff Code \\
		8 & RESFLG & Resident Flag \\
		9 & BKGSINCEDT & Date since when banking with the Bank \\
		10 & GENCD & Gender Code \\
		11 & BIRTHDT & Date of Birth \\
		12 & MARST & Marital Status \\
		13 & OCUCD & Occupation Code \\
		14 & DRYLAND & Area of DRY Land \\
		15 & WETLAND & Area of WET Land \\
		16 & SBWCLMTAMT & SBI Working Capital Limit \\
		17 & PARTBKFLG & Participating Bank Flag \\
		18 & SBTLLMTAMT & SBI Term Loan Limit \\
		19 & OPINIONDT & Opinion Date \\
		20 & TPGAMT & Third Party Guarantee Amount \\
		21 & BORWORAMT & Net Worth of the Borrower \\
		22 & PRODCODE & Product Code \\
		23 & FACCD & Facility Code \\
		24 & SUBFACCD & Sub Facility Code \\
		25 & PRIFLG & Priority Flag \\
		26 & DIRFINFLG & Direct Finance Flag \\
		27 & SECTORCD & Sector Code \\
		28 & SCHEMECD & Scheme Code \\
		29 & ACTCD & Account ID \\
		30 & SANCTIONDT & Date of Sanction \\
		31 & SANAUTCD & Sanctioning Authority \\
		32 & OPENINGDT & Date of Opening \\
		33 & LIMITAMT & Limit Amount \\
		34 & DOCREVDT & Document Revival Date \\
		35 & FOODNONFLG & Food or Non-Food Sector Flag \\
		36 & INTRATE & Rate of Interest \\
		37 & OUTAMT & Amount Outstanding \\
		38 & PRISECCD1 & Primary Security Code 1 \\
		39 & PRISECCD2 & Primary Security Code 2 \\
		40 & PRISECAMT & Primary Security Amount \\
		41 & SPLCSAMT & Specific Collateral Security Amount \\
		42 & INCAMT & Interest Not Collected Amount \\
		43 & DISTTCD & District Code \\
		44 & POPCD & Population Code \\
		45 & UNIFUNFLG & Unit Function Status Flag \\
		46 & ACCSTACD & Accounting Std Code - Previous Quarter \\
		47 & FOODNONFLG & Food or Non-Food Sector Flag \\
		48 & RENEWALDT & Date of Renewal \\
		49 & ALLCUSTID & Customer ID \\
		50 & DPAMT & Drawing Power Amount \\
		51 & INSEXPDT & Date of Insurance Expiry \\
		52 & INSAMT & Insurance Amount (Sum Assured) \\
		53 & FULDISFLG & Loan Fully Disbursed? \\
		54 & LASCREDT & Date of Last Credit \\
		55 & REPTYPCD & Repayment Type Code \\
		56 & PERIOD & Frequency of Interest Application \\
		57 & YTDINTAMT & Year-to-date interest applied \\
		58 & YTDCSUMAMT & Year-to-date credit summation \\
		59 & QDISMADAMT & YTD Disbursement (i.e. During the Year) \\
		60 & NUMOFINST & No of Instalments \\
		61 & REPFRQCD & Repayment Frequency Code \\
		62 & FIRINSDT & Date of First Instalment \\
		63 & INSTALAMT & Instalment Amount \\
		64 & RECAMT & INCA Recovered up to the Quarter \\
		65 & CALC\_ASSET & Calculated Asset (CALC) \\
		66 & TFRDT & Date of Transfer to Recalled Assets \\
		67 & REASONCD & Reason for Transfer to Recalled Assets \\
		68 & GINAMT & Amount originally Transferred to RA \\
		69 & RECALLDT & Date of Recall \\
		70 & SUTFILAMT & Suit Filed Amount \\
		71 & CUSTTYPE & Customer Type \\
		72 & RETAINAMT & DICGC/ECGC/CGTSI claim Retainable Amount \\
		73 & WOSACD & Retaining Amount \\
		74 & CUSTOTLMT & Customer Total Limit (CALC) \\
		75 & CUSTOTOUT & Customer Total Outstanding (CALC) \\
		76 & SUBSIDYAMT & Subsidy received and held amount \\
		77 & NFMRGAMT & Out of Total Security, Cash Security \\
		78 & MAR\_CALC & Calculated Asset as on March (Last) \\
		79 & MAR\_URIPY & URIPY as on March (Last) \\
		80 & PROV\_TOTAL & Current Provision Amount \\
		81 & IRAC & 1 - Std, 2 Sub-Std, 3-Doubtful, 4 Loss \\
	\end{longtable}
	
\section*{Data Availability Statement}
The datasets analyzed during the current study are not available due to consumer privacy issues.

\section*{Author Information}

Authors and Affiliations:\\
\textbf{Fellow of Decision Science}, IIM Mumbai, India.\\
\href{mailto:minati@example.com}{Minati Rath}.\\
\textbf{Faculty of Decision Science}, IIM Mumbai, India.\\
\href{mailto:hemadate@iimmumbai.ac.in}{Hema Date}.

\section*{Corresponding Authors}
Correspondence to \href{mailto:minati06@gmail.com}{Minati Rath} or \href{mailto:hemadate@iimmumbai.com}{Hema Date}.

\end{document}